\documentclass[letterpaper, 10 pt, conference]{ieeeconf}  

\IEEEoverridecommandlockouts                              

\usepackage{graphicx} 
\usepackage{xcolor}
\usepackage{gensymb}

\usepackage{multirow}
\usepackage[linesnumbered,ruled,vlined]{algorithm2e}
\usepackage{textcomp}
\usepackage{xcolor}
\usepackage{amsmath,amssymb,amsfonts}
\usepackage{indentfirst}
\usepackage{enumerate}
\usepackage{float}
\usepackage[subrefformat=parens,labelformat=parens, caption = false]{subfig}

\usepackage[sorting=none, style=ieee]{biblatex}

\graphicspath{{./Figure/}}
\title{\LARGE \bf
Learning Pseudo Front Depth for 2D Forward-Looking Sonar-based Multi-view Stereo
}

\author{Yusheng Wang$^{1}$, Yonghoon Ji$^{2}$, Hiroshi Tsuchiya$^{3}$, Hajime Asama$^{1}$, and Atsushi Yamashita$^{4}$
\thanks{
This work was partly supported by JSPS KAKENHI Grant Number 21J12362.}
\thanks{$^{1}$Y.~Wang, H.~Asama are with the Department of Precision Engineering, Graduate School of Engineering, The University of Tokyo, Japan.~{\tt\small \{wang,asama\}@robot.t.u-tokyo.ac.jp}}
\thanks{$^{2}$Y. Ji is with the Graduate School for Advanced Science and Technology, JAIST, Japan. {\tt\small ji-y@jaist.ac.jp}}
\thanks{$^{3}$H. Tsuchiya is with the Research Institute, Wakachiku Construction Co., Ltd., Japan. {\tt\small hiroshi.tsuchiya@wakachiku.co.jp}}
\thanks{$^{4}$A.~Yamashita is with the Department of Human and Environmental Engineered Studies, Graduate School of Frontier Sciences, The University of Tokyo, Japan.~{\tt\small yamashita@robot.t.u-tokyo.ac.jp}}
}

\begin{document}

\maketitle
\thispagestyle{empty}
\pagestyle{empty}

\begin{abstract}
Retrieving the missing dimension information in acoustic images from 2D forward-looking sonar is a well-known problem in the field of underwater robotics. 
There are works attempting to retrieve 3D information from a single image which allows the robot to generate 3D maps with fly-through motion. However, owing to the unique image formulation principle, estimating 3D information from a single image faces severe ambiguity problems. Classical methods of multi-view stereo can avoid the ambiguity problems, but may require a large number of viewpoints to generate an accurate model. 
In this work, we propose a novel learning-based multi-view stereo method to estimate 3D information. To better utilize the information from multiple frames, an elevation plane sweeping method is proposed to generate the depth-azimuth-elevation cost volume. The volume after regularization can be considered as a probabilistic volumetric representation of the target. Instead of performing regression on the elevation angles, we use pseudo front depth from the cost volume to represent the 3D information which can avoid the 2D-3D problem in acoustic imaging.  
High-accuracy results can be generated with only two or three images. Synthetic datasets were generated to simulate various underwater targets. We also built the first real dataset with accurate ground truth in a large scale water tank. Experimental results demonstrate the superiority of our method, compared to other state-of-the-art methods.

\end{abstract}

\section{INTRODUCTION}
When water visibility is poor, sonar is usually the only viable modality for underwater sensing. High-frequency 2D forward-looking sonar, or acoustic cameras, can generate high-resolution 2D acoustic images in any body of water. They are compact in size and can be easily mounted on remotely operated vehicles (ROVs) or autonomous underwater vehicles (AUVs). In the last few years, the sensors have achieved remarkable results in tasks such as mosaicking, 3D mapping, and robot navigation \cite{Hurtos2015,Wangjoe,Wangicra2021,Negahdaripour2013}. Because sonar is a 2D imaging sensor, dimension missing problem also exists similar to optical cameras. Unlike the optical camera where the depth information is missing, the information in the elevation angle direction is missing during the image formation for acoustic cameras. Retrieving 3D information from 2D acoustic images is a nontrivial problem. The unique imaging model, severe noise, and sonar artifacts make this problem difficult.

Early works on 2D forward-looking sonar utilized sparse features, such as corners or edges, for 3D reconstruction \cite{Mai20171,Mai20172,Huang2015}. Points or lines were used to represent the sparse 3D model, which is not intuitive for human comprehension. 
More recently, dense 3D reconstruction methods have been proposed, based on multi-view stereo or photometric stereo. Multi-view stereo methods basically utilize an efficient space carving scheme, considering the known sensing range and limited field of view (FoV) of 2D forward-looking sonars \cite{Aykin2017,guerneve2018,Wangjoe}. However, the deterministic multi-view stereo methods require a large number of viewpoints; the 3D reconstruction result is extremely poor when only a few viewpoints (e.g., 2$\sim$3) are available. Photometric stereo methods can generate a 3D model from a single acoustic image by modeling ultrasound propagation or using shadow information \cite{Aykin2016,Westman2019iros}. However, such methods require ideal conditions, and usually have strong assumptions about the scene. 
Recently, learning-based methods have demonstrated successful results in computer vision problems. A convolutional neural network (CNN)-based method was proposed to estimate the elevation angle for each pixel from a single acoustic image \cite{DeBortoli2019}. However, for acoustic images, one pixel may correspond to multiple elevation angles, which is a \textit{non-bijective 2D-3D correspondence problem}, denoted as 2D-3D problem in the following paper. To solve this problem, our research group proposed a CNN-based method (A2FNet) to learn the front view depth from a single acoustic image \cite{Wangicra2021}. 

In fact, estimating 3D information from a single acoustic image is a highly ill-posed problem and cannot be solved for many cases, considering the ambiguity existing in acoustic imaging (see Section III.C). 
In addition, the accuracy of single view 3D reconstruction is limited. Recently, studies have focused on sonar stereo \cite{Negahdaripour2020,McConnell2020}. Previous studies on sonar stereo still require handcrafted features that may not perform stably. In this work, we also emphasize the benefits of sonar stereo. Deterministic multi-view stereo methods can solve the ambiguity problem caused by single-view 3D reconstruction, but may require a large number of viewpoints. This paper aims to solve the 3D reconstruction problem by multi-view stereo but use only 2$\sim$3 viewpoints. 

\begin{figure*}[tb]
\centering
{\includegraphics[width=1.9\columnwidth]{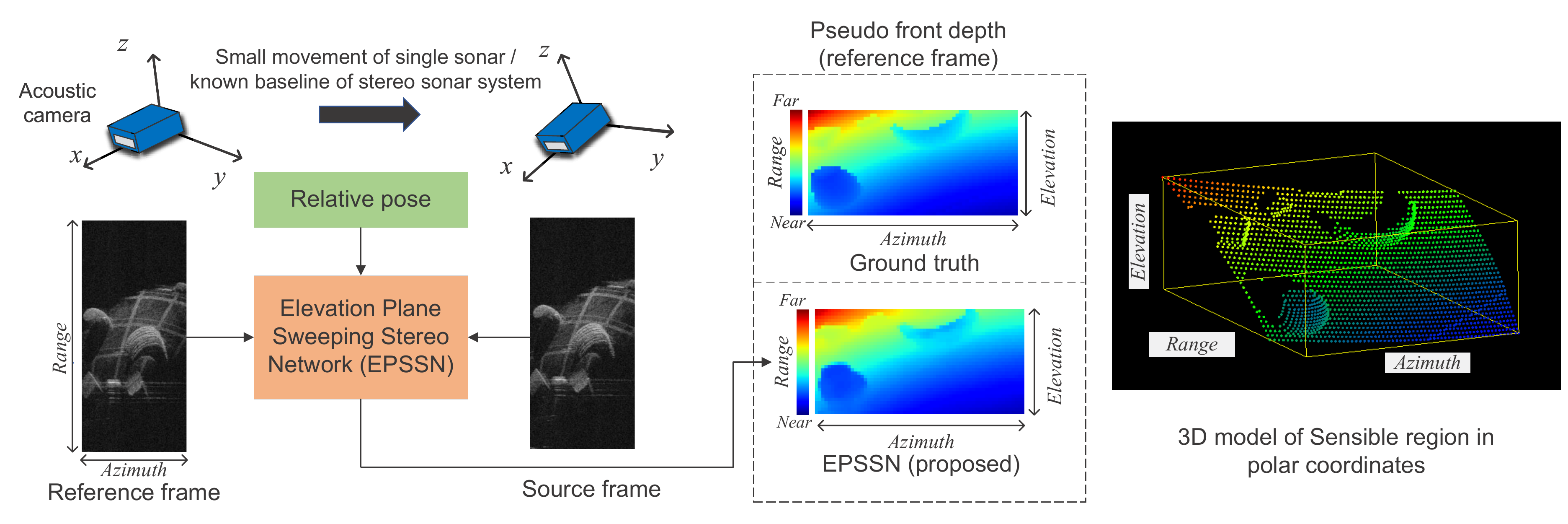}}
\caption{Workflow of the proposed method. The reference frame and the source frames of sonar stereo are input to the elevation plane sweeping stereo network (EPSSN) with the corresponding relative poses. The EPSSN estimates the pseudo front depth of the reference frame. The acoustic image is an azimuth-range image in which the elevation information is missing. The 3D information can be retrieved by generating azimuth-elevation depth map. } 
\label{fig:catch}
\end{figure*}

Retrieving 3D information using only a few viewpoints is still difficult, in this work, we propose a deep learning based method for the multi-view sonar stereo problem as shown in Fig.~\ref{fig:catch}. We propose a novel \underline{E}levation \underline{P}lane \underline{S}weeping \underline{S}tereo \underline{N}etwork (EPSSN). The major task is to retrieve the 3D information of the reference frame. The source frames are the ones from viewpoints with small motion change to help the 3D reconstruction of the reference frame. The relative pose between the reference frame and the source frames are assumed given in this work, which can be acquired from control input of a motor, inertial measurement unit (IMU) or a known baseline of sonar stereo system. 
By providing reference and source frames with relative pose information, it is possible to generate a highly accurate pseudo front view depth for the reference frame. By warping the features of the source frame to virtual elevation planes in the reference frame, it is possible to generate a depth-azimuth-elevation cost volume. Finally, the front view depth can be directly generated from the cost volume according to the geometry relationship. It is noticed that not only the ambiguity can be solved by using 2$\sim$3 viewpoints instead of a single viewpoint, the accuracy of 3D reconstruction results may increase drastically. 

Our contributions are listed as follows.
\begin{itemize}
    \item We propose a novel CNN-based method to generate a dense 3D reconstruction from few (e.g., two) acoustic images.
    \item We propose an elevation plane sweeping scheme to generate the depth-azimuth-elevation cost volume. The front view depth can be calculated from the cost volume. By using the front view depth, the 2D-3D problem can be avoided.
    \item We generate synthetic and real datasets for evaluation. The synthetic datasets include natural and artificial scenes, and have been released under an open-source license\footnote{https://github.com/sollynoay/EPSSN}. Our proposed method outperforms other state-of-the-art methods. 
\end{itemize}


\section{Related Works}

\subsection{3D Reconstruction of Acoustic Images}
Early methods used sparse points or lines for the 3D reconstruction of acoustic images. Mai et al. and Huang et al. used manually selected corner points combined with simultaneous localization and mapping (SLAM) systems for 3D information acquisition \cite{Mai20171,Huang2015}. For automatic feature detection, AKAZE features have been proven to be effective in acoustic images that have been used for sonar-based localization \cite{Li2018}. Wang et al. tracked AKAZE features based on optical flow and assumed the terrain as a Gaussian process random field on a tree structure \cite{wangj2019}. The terrain is assumed to be smooth and the computation cost increases drastically when the feature points increase. 

For better 3D representation, dense 3D reconstruction has been studied. For multi-view stereo methods, Aykin et al. applied a space carving method for small objects \cite{Aykin2017}. The acoustic images were binary-segmented into object and non-object regions. The space was then voxelized and projected onto the binary image. The voxels projected to the object region were kept and the others were carved. Guerneve et al. utilized a linearized sonar projection model, with min-filtering for space carving \cite{guerneve2018}. Our group proposed a probabilistic method using 3D occupancy mapping with an inverse sensor model \cite{Wangjoe}. It can be applied to more general scenes such as 3D mapping. However, the aforementioned methods require a large number of viewpoints and do not work well with only one or two viewpoints. Photometric stereo methods utilize shadows or other image cues to retrieve 3D information. Aykin et al. physically modeled ultrasound propagation and scenes, and proved that the physical model can be used for 3D reconstruction with object contours \cite{Aykin2016}. Westman et al. utilized a similar scheme and generated 3D model of continuous surface \cite{Westman2019iros}. Non-line-of-sight (NLOS) methods have also been applied to acoustic images and proved to be feasible \cite{Westman2020iros}. However, these methods usually have strong assumptions, e.g., that shadows or Fermat paths \cite{Westman2020iros} can be accurately detected, which is difficult, owing to noise and artifacts. 

Recent deep learning-based methods have also been utilized in this field. Debortoli et al. used a UNet-like structure for pixel-wise elevation angle estimation of a single acoustic image \cite{DeBortoli2019}. For the synthetic dataset, the network was trained in a supervised manner. For the real dataset, the network was first trained using synthetic images and fine-tuned using a self-supervised scheme with small motions. The study assumes that each pixel only corresponds to one elevation angle which is not always true. To overcome this limitation, our group proposed a network to learn the front view depth, which is a better representation with no 2D-3D ambiguity problem \cite{Wangicra2021}. The network implicitly learns to transfer an azimuth-range image to an azimuth-elevation depth map. Retrieving 3D information from a single acoustic image is highly ill-posed, and cannot be solved, owing to the ambiguity in many cases that are difficult to apply in real underwater scenarios. 

Negahdaripour first proposed sonar stereo, established sonar epipolar geometry theory and proved the optimal configuration of two forward-looking sonars \cite{Negahdaripour2020}. McConnell et al. mounted two orthogonally configured sonars to an ROV to generate dense point clouds \cite{McConnell2020}. Such methods still require handcrafted features in the acoustic images. In this work, we are the first to propose a learning-based method for multi-view sonar stereo. It can be used for structured sonar stereo and multiple images from a single sonar. It can solve the problems caused by sonar imaging and generate accurate dense 3D reconstruction results even for complex underwater scenarios. 

\subsection{Learning Multi-view Stereo}

For optical cameras, retrieving 3D information from stereo cameras is a long-standing problem. Recently, learning-based end-to-end depth from stereo matching methods has shown outstanding abilities \cite{zbontar2016,Mayer2016,Kendall2017}. By building 3D cost volumes to fuse the information from stereo images, and using soft argmax to estimate the depth of each pixel, the neural networks show high performance for depth estimation problems \cite{Kendall2017}. However, they usually assume that the two optical cameras are well  calibrated and the image pairs are rectified. 

Recently, studies have been conducted on unstructured multi-view stereo \cite{mvsnet,deepmvs}. To build the cost volume, plane sweeping scheme \cite{Collins1996} is applied, by projecting images or features to virtual depth planes in the reference camera coordinates. The projection process was calculated using homography. After acquiring the cost volume, multi-view stereo usually follows the same structure as stereo matching methods. Acoustic cameras have a different projection theory compared to optical cameras, so theories such as homography cannot be applied. In addition, the depth estimation of optical images directly estimates the depth of each pixel in the optical image. However, instead of estimating the elevation angles for each pixel, we estimate the pseudo front depth from the acoustic images.    

\section{Preliminaries}

\subsection{Sonar Projection Theory}

\begin{figure}[tb]
\centering
{\includegraphics[width=1.0\columnwidth]{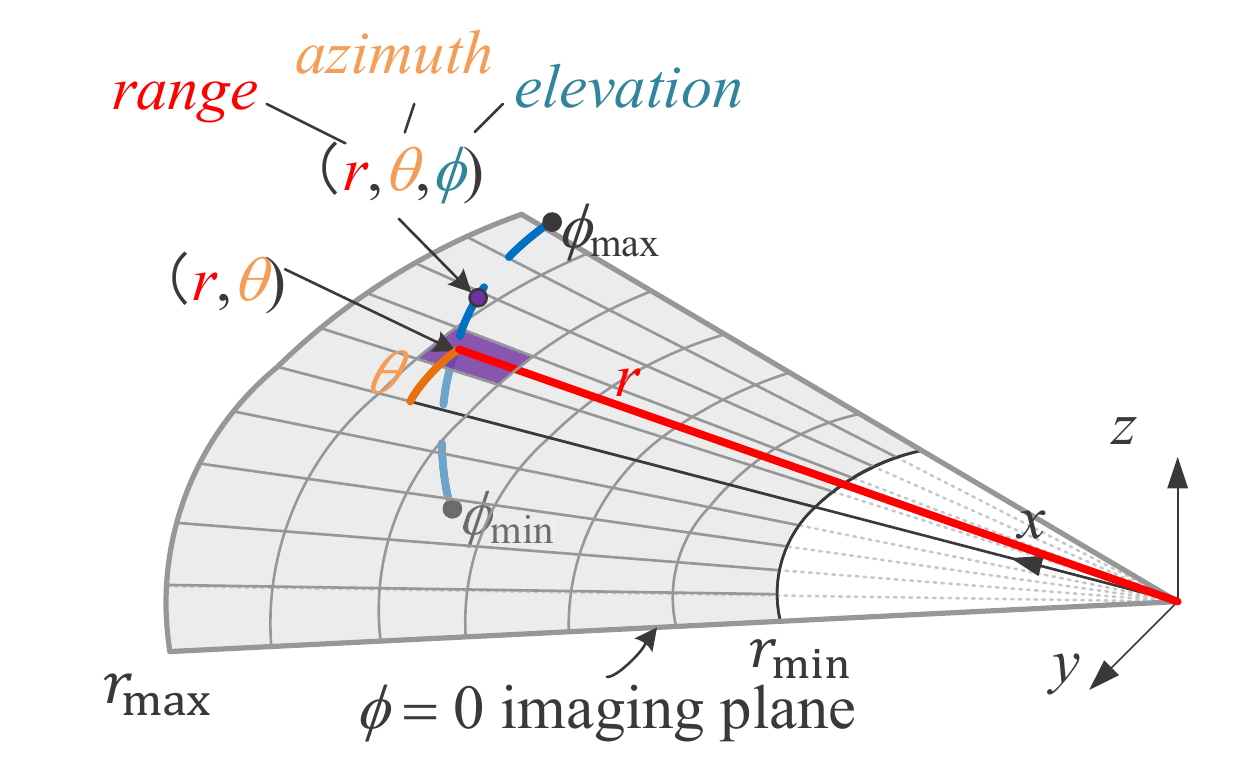}}
\caption{Geometry model of 2D forward looking sonar. It can be equivalently seen as a projection of a 3D point to the $\phi=0$ plane, where the elevation information is missing. } 
\label{fig:projection}
\end{figure}

A 3D point in the sonar coordinate system is usually represented as $(r,\theta,\phi)$ in the polar coordinate system, which can be transformed to the Euclidean coordinate as follows. 

\begin{equation}
	\begin{bmatrix}
	X_c & Y_c & Z_c 
	\end{bmatrix}^{\top}
	=
	\begin{bmatrix}
	r\cos \phi \cos \theta & r \cos \phi \sin \theta & r \sin \phi 
	\end{bmatrix}^{\top}.
\end{equation}

During the imaging process, the $\phi$ angle information is missing, and the corresponding 2D point in the image coordinates can be written as $(r,\theta)$ as shown in Fig.~\ref{fig:projection}. Different from optical imaging, each pixel in the acoustic images may correspond to multiple 3D points. The backscattered intensity of the 3D points with the same $(r,\theta)$ accumulates. For each $(r,\theta)$ in the acoustic image $I_a$, discretizing $\phi$ as $\{\phi_1,\phi_2,\dots,\phi_n\}$ and the $i$-th corresponding backscattered intensities as $I(r,\theta,\phi_i)$ gives 
$
I_a(r,\theta) = \sum_{i=1}^{n}I(r,\theta,\phi_i).
$
\subsection{Pseudo Front Depth} 
Considering the ultrasound as rays, each ray strikes the object surface and reflected to sensor. Before integration in the elevation angle direction, it can be seen as a common time-of-flight (ToF) sensor, from which a depth map can be acquired. Theoretically, a front depth map contains the entire 3D information of the sensible region as shown in Fig.~\ref{fig:catch}. Instead of estimating the elevation angles directly for each pixel, the front depth map representation can avoid the 2D-3D problem \cite{Wangicra2021}. However, it is less constrained than the direct elevation angle estimation problem. 
\subsection{Ambiguity Problem}
\begin{figure}[tb]
\centering
{\includegraphics[width=1.0\columnwidth]{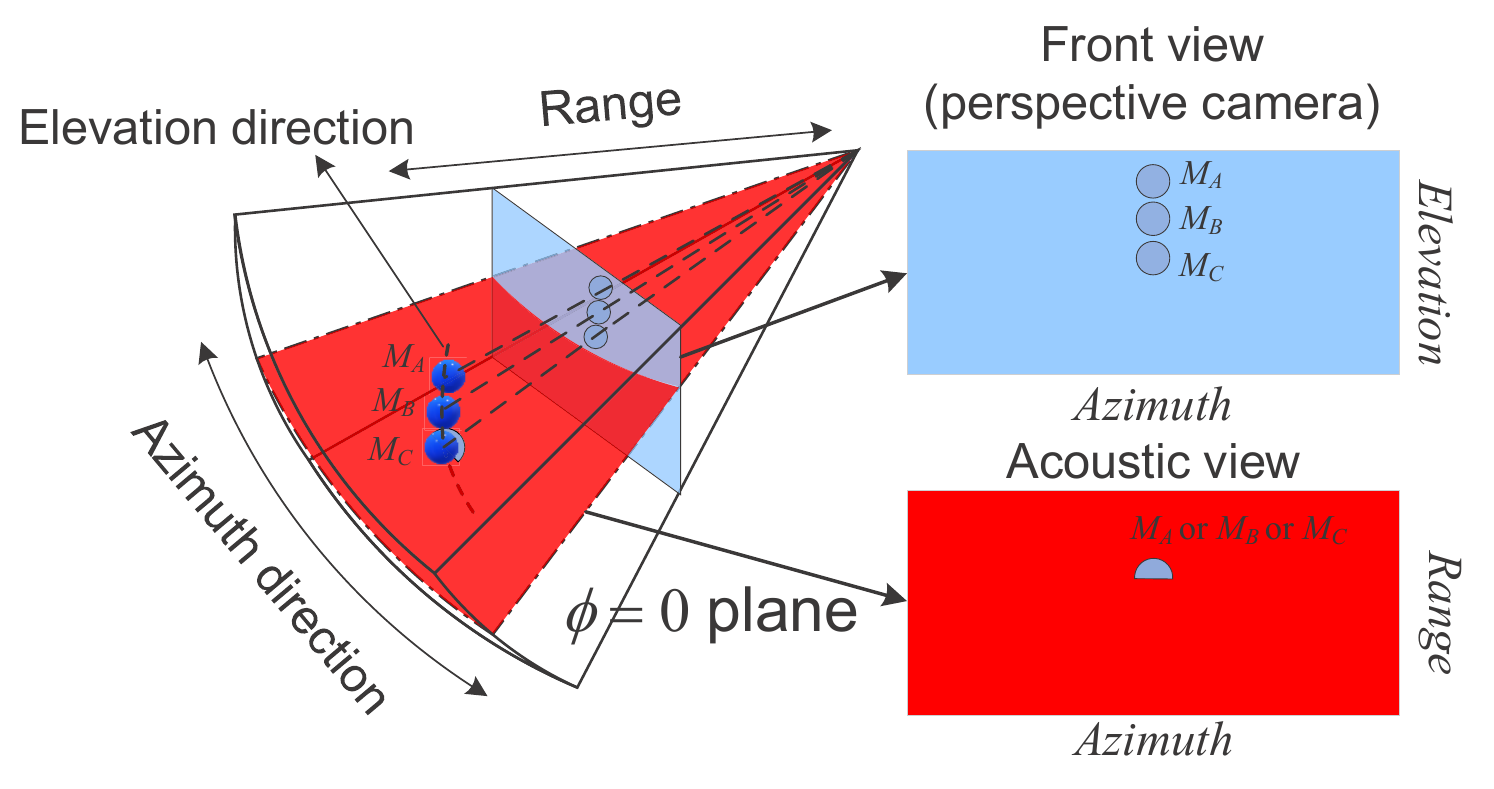}}
\caption{An example of the ambiguity problem. The spheres from ${\rm M}_A$ to ${\rm M}_C$ may generate the same acoustic image. This may cause ambiguity for inverse problem.} 
\label{fig:ambiguity}
\end{figure}

Sonar imaging faces severe ambiguity problems. For example, when sensing a single spherical object, theoretically, all spheres from ${\rm M}_A$ to ${\rm M}_C$ will generate the same acoustic image as shown in Fig.~\ref{fig:ambiguity}. Estimating the sphere position from a single image is highly ill-posed. In addition, objects or scenes symmetric to the imaging plane will generate the same image. 
For CNN-based regression methods, if one input corresponds to multiple ground truth labels, the network cannot learn the problem well. Therefore, it is necessary to study sonar stereo to overcome the aforementioned problems. 

\section{Learning Sonar Stereo}

\subsection{Overview}

\begin{figure*}[tb]
\centering
{\includegraphics[width=1.9\columnwidth]{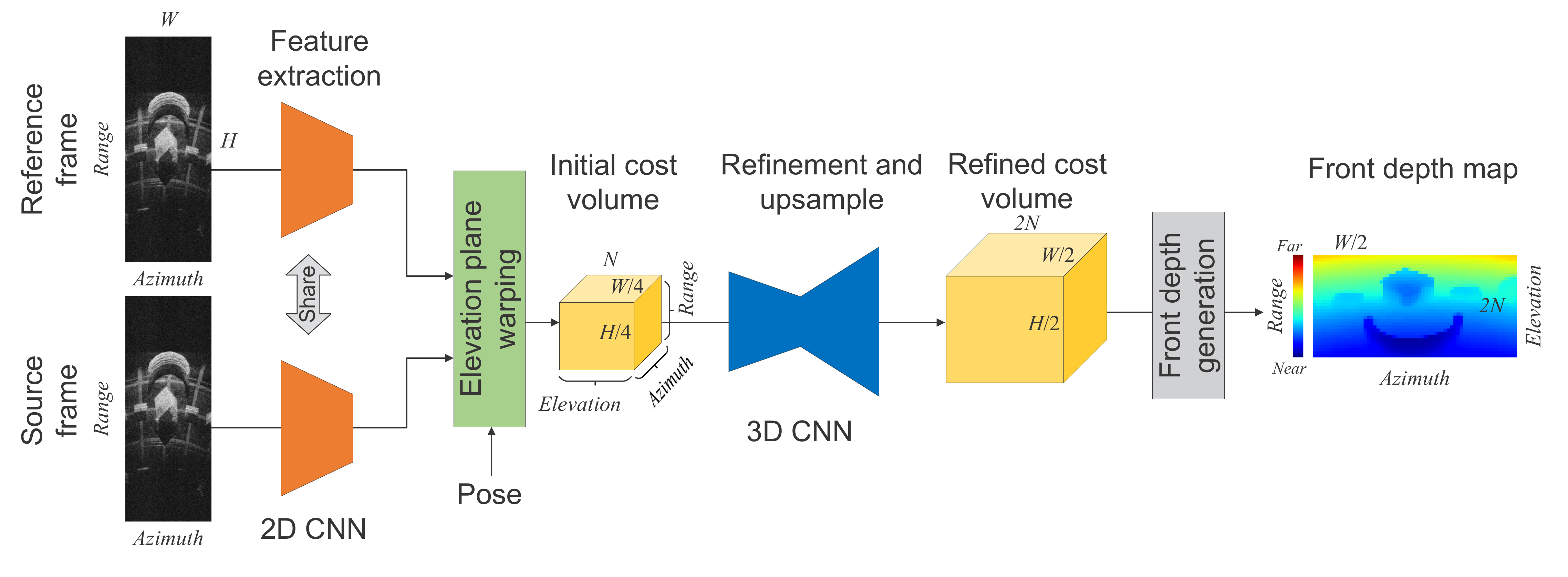}}
\caption{Overview of EPSSN. The reference and source frames are passed through a 2D CNN for feature extraction. We propose a novel feature-level elevation plane warping method to generate the depth-azimuth-elevation cost volume. A 3D CNN is used to refine and upsample the cost volume. Then, a front depth generation block is proposed to generate the front depth map. } 
\label{fig:structure}
\end{figure*}

Figure~\ref{fig:structure} shows the network structure overview of EPSSN. The inputs are the reference frame and the source frames with corresponding poses. The output is the pseudo front depth map of the reference frame. We propose two major novel blocks: the elevation plane warping and front depth generation blocks. 
A 2D CNN is first used to extract deep features $\bf{F}$ from the acoustic images. The features in the source frames are warped to the reference frame based on elevation plane sweeping to generate cost volumes as explained in Section IV.B. In this study, we use a depth-azimuth-elevation volume. After fusing the cost volumes from the reference frame and the source frames, a following 3D CNN is used to refine and upsample the initial cost volume. Finally, the front depth generation block is used to calculate the front depth map from the cost volume, as explained in Section~IV.C. 


\subsection{Elevation Plane Warping}

\begin{figure}[tb]
\centering
{\includegraphics[width=1.0\columnwidth]{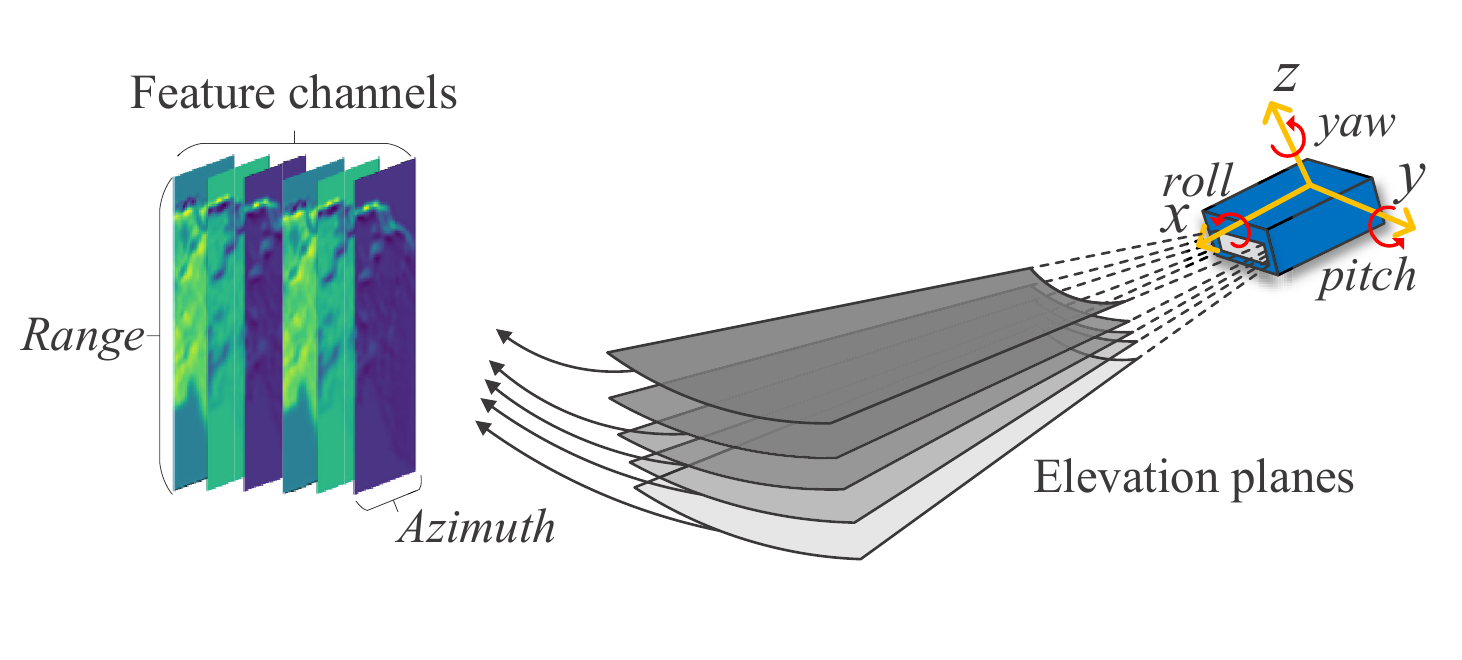}}
\caption{Elevation plane warping. We generate virtual elevation planes along elevation angle direction. Each plane warps the deep feature and all the warped features are concatenated in a new dimension which refers to the elevation direction.} 
\label{fig:warp}
\end{figure}

Plane sweeping stereo has been commonly used in multi-view stereo for optical cameras \cite{Collins1996}. Multi-view images are projected onto virtual depth planes in the reference image coordinates to generate a cost volume. The depth maps are then estimated using this cost volume. This type of theory has never been used for acoustic images because of the difference in projection theory. In this work, we propose an elevation plane sweeping scheme for acoustic images to generate a depth-azimuth-elevation volume and use the volume to further generate front depth maps. We first generate $N$ virtual elevation planes along the elevation angle direction as shown in Fig.~\ref{fig:warp}. The elevation angle $\phi$ of the $j$-th plane can be described as follows.
\begin{equation}
    {\phi}_j=\phi_{\rm min}+\frac{j(\phi_{\rm max}-\phi_{\rm min})}{(N-1)},
\end{equation}
where $j$ is an integer from $0$ to $N-1$. 
For each virtual elevation plane, denoting a 3D point $i$ in plane $j$ as ${\bf p}_{i,j}^{src}=({r}_{i,j},{\theta}_{i,j},{\phi}_j)$, the point is first transformed into the Euclidean coordinate as ${\bf e}_{i,j}^{src}$. The point in the reference frame is then transformed into the source coordinate as ${\bf e}_{i,j}^{ref}={\bf T}_{ref}^{src}{\bf e}_{i,j}^{src}$. Here, ${\bf T}_{ref}^{src}$ denotes the relative motion between the reference and source frames. Then, we transform the point to the polar coordinate as ${\bf p}_{i,j}^{ref}$ and project the point onto the imaging plane in the source frame. We use bilinear sampling to acquire the corresponding feature $\bf{F}$ in source frame. Then, we can warp feature ${\bf F}$ to elevation plane $j$ and the result is denoted as ${\bf F}_j^{wp}$. The warped source frame cost volume ${\bf V}_{src}$ can be generated as follows. 
\begin{equation}
    {\bf V}_{src} = CAT({\bf F}_0^{wp},\dots,{\bf F}_{N-1}^{wp}),
\end{equation}
where $CAT$(.) refers to the concatenation operation along a new dimension corresponding to the elevation direction. The initial cost volume is a 4D tensor, for which the shape is $C\times N \times H/4 \times W/4$, where $C$ refers to the length of feature channels. The cost volume for the reference frame itself is denoted as ${\bf V}_{ref}$. The volumes are aggregated and the initial cost volume $\bf{C}$ is calculated as follows.
\begin{equation}
    {\bf C} = \frac{({\bf V}_{ref}-\bar{\bf V})^2+({\bf V}_{src}-\bar{\bf V})^2}{\rm 2},
\end{equation}
where $\bar{\bf V}$ is the element-wise arithmetic mean of ${\bf V}_{src}$ and ${\bf V}_{ref}$. This can be easily extended if there are multiple source frames.


\subsection{Front Depth Generation}

\begin{figure}[tb]
\centering
{\includegraphics[width=1.0\columnwidth]{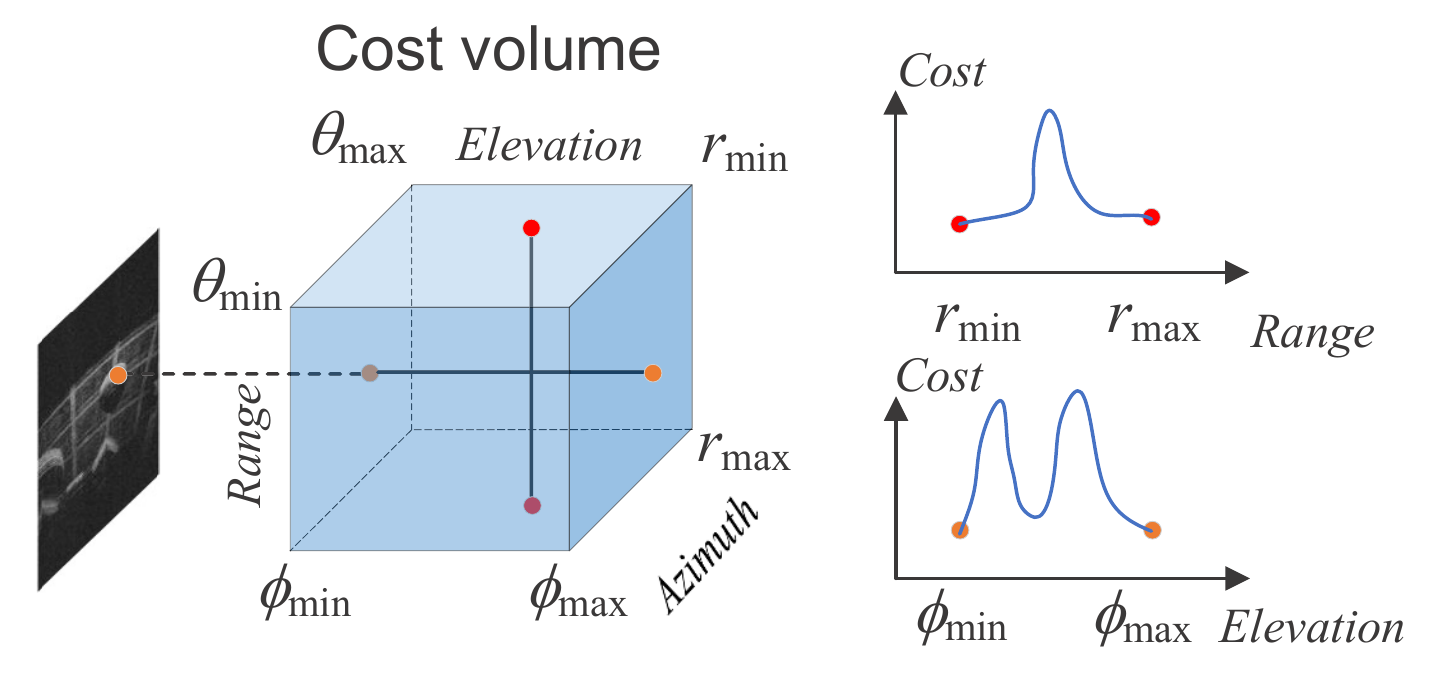}}
\caption{Refined cost volume. The refined cost volume can be considered as a probabilistic volumetric representation of the target. In our problem, the probability distribution along the elevation direction may be multi-modal. On the other hand, we assume that the distribution along the range direction is always uni-modal. } 
\label{fig:cost}
\end{figure}

The initial cost volume is noisy and not regularized. A 3D UNet-like structure \cite{unet} is used to refine the initial cost volume, followed by an extra 3D convolution transpose layer to further upsample the cost volume. The last convolution layer makes the size of feature channels to one. 

The refined cost volume can be considered as a probabilistic volumetric representation of the target as shown in Fig.~\ref{fig:cost}. Each voxel has a probability value of occupancy and corresponds to the 3D space in the reference coordinate. In particular, in our cases, the boundaries of the cost volume refer to the scope of the sonar. In this work, we assume that the front depth representation is ideal so that the probability distribution for the range of a $(\theta,\phi)$ pair is always unimodal. On the other hand, the probability distribution for the elevation angle of a $(r,\theta)$ pair is more likely to be multimodal due to the non-bijective projection as explained in Section~III.A. Theoretically, the front depth map can be retrieved by winner-take-all (i.e., argmax) \cite{Kendall2017}. However, it is difficult to use the same scheme for the multi-modal distribution of the elevation map case. Because argmax cannot offer subpixel results and is not differentiable for network training, it is usually replaced by soft argmax in the neural network. The depth $\hat{\bf{D}}$ for $(\theta,\phi)$ is calculated as follows using softmax and the depth value for each cell.
\begin{equation}
    \hat{{\bf D}}(\theta,\phi) = \frac{1}{\sum_{j=r_{\rm min}}^{r_{\rm max}}e^{{\bf C}(j,\theta,\phi)}}\sum_{d=r_{\rm min}}^{r_{\rm max}}d\times e^{{\bf C}(d,\theta,\phi)}.
\end{equation}

\subsection{Loss Functions}

In this study, we estimate the depth information instead of reverse depth. This is mainly because the sensing range $|r_{\rm max}$-$r_{\rm min }|$ is not a large value (e.g. 2~m) in this work. This may lose precision, owing to float-point numbers when using reverse depth. The shortcoming of using direct depth is that some of the values in the ground truth depth value may exceed $r_{\rm max}$, especially for submerging objects, and some depth values may be nearly infinite. During the training time, we provide a binary mask $\mathcal{M}$ from depth labels ${\bf{D}}$ as follows. 
\begin{equation}
    \mathcal{M}(\theta,\phi)=\left \{
    \begin{array}{rcl}
      0 &   & {\bf D}(\theta,\phi)\leq r_{\rm min} \lor {\bf D}(\theta,\phi)\geq r_{\rm max}  \\
      1 &   & r_{\rm min}<{\bf D}(\theta,\phi)<r_{\rm max}
    \end{array}.
    \right.
\end{equation}
The loss function is written as follows. 
\begin{equation}
    \mathcal{L} = \frac{1}{n}\sum_{i=1}^n{\lambda}{\mathcal{M}_i}|\hat{\bf D_i}-{\bf D_i}|+{\bar{\mathcal{M}}_i}|\hat{\bf D_i}-{\bf D_i}|,
\end{equation}
where $\bar{\mathcal{M}}$ refers to $1-{\mathcal{M}}$ and $\lambda$ is a hyperparameter that balances the two terms in the equation. Noting that for ground truth depth maps, during training, we set the range larger than $r_{\rm max}$ to $r_{\rm max}$ and the range smaller than $r_{\rm min}$ to $r_{\rm min}$. We assume that the object or scene targets are away from the boundaries. It is then easy to filter the non-informative estimations during the test times.

\section{Experiment}

To evaluate the proposed methods, we generated both simulation and real datasets. 
Usually, roll rotation is considered to be efficient for retrieving 3D information. Existing sonar stereo systems use concurrent orthogonal sonars for sensing which can be approximately considered as a roll rotation of 90$^{\circ}$ \cite{Negahdaripour2020,McConnell2020}. However, to ensure a sufficient overlap of the images, we used known small roll rotations (5$^\circ\sim$10$^\circ$) to generate image pairs in this work. The influence of motion will be investigated in future studies. 

\subsection{Simulation Datasets}
The simulation datasets were generated in Blender environment which can be found on our GitHub\footnote{https://github.com/sollynoay/Sonar-simulator-blender}. We prepared three datasets with different scenes, including artificial environment, terrain, and sphere. The artificial environment was built using a 3D CAD model which is also the ground truth of our real dataset. Image pairs were generated at random positions throughout the scene. We generated 4,000 image pairs for training and 2,000 pairs for test. For the terrain dataset, we generated terrains with the A.N.T. Landscape add-on in Blender by using hetero noise, which simulated undulating seabed. We trained the network on one terrain and tested it on another terrain. We generated 1,000 image pairs for training and 1,000 pairs for test. The sphere dataset was used to verify the ambiguity problem explained in Section III.C. To generate the dataset, the sphere was located at a random position within the sonar scope. 

\subsection{Real Datasets}

\begin{figure}[tb]
\centering
\subfloat[ \label{device}]{\includegraphics[width=0.475\columnwidth]{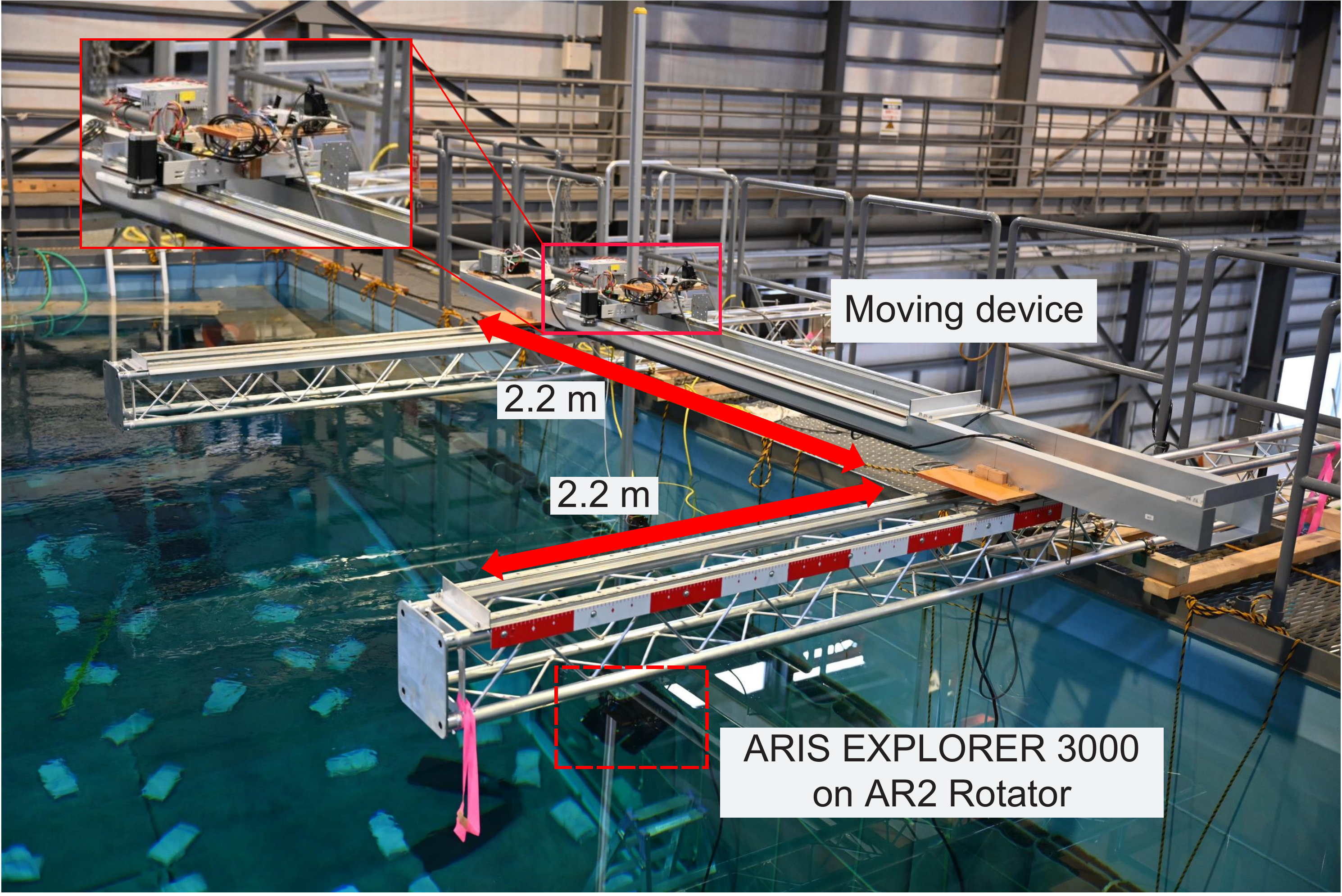}}\enskip
\subfloat[ \label{realtarget}]{\includegraphics[width=0.475\columnwidth]{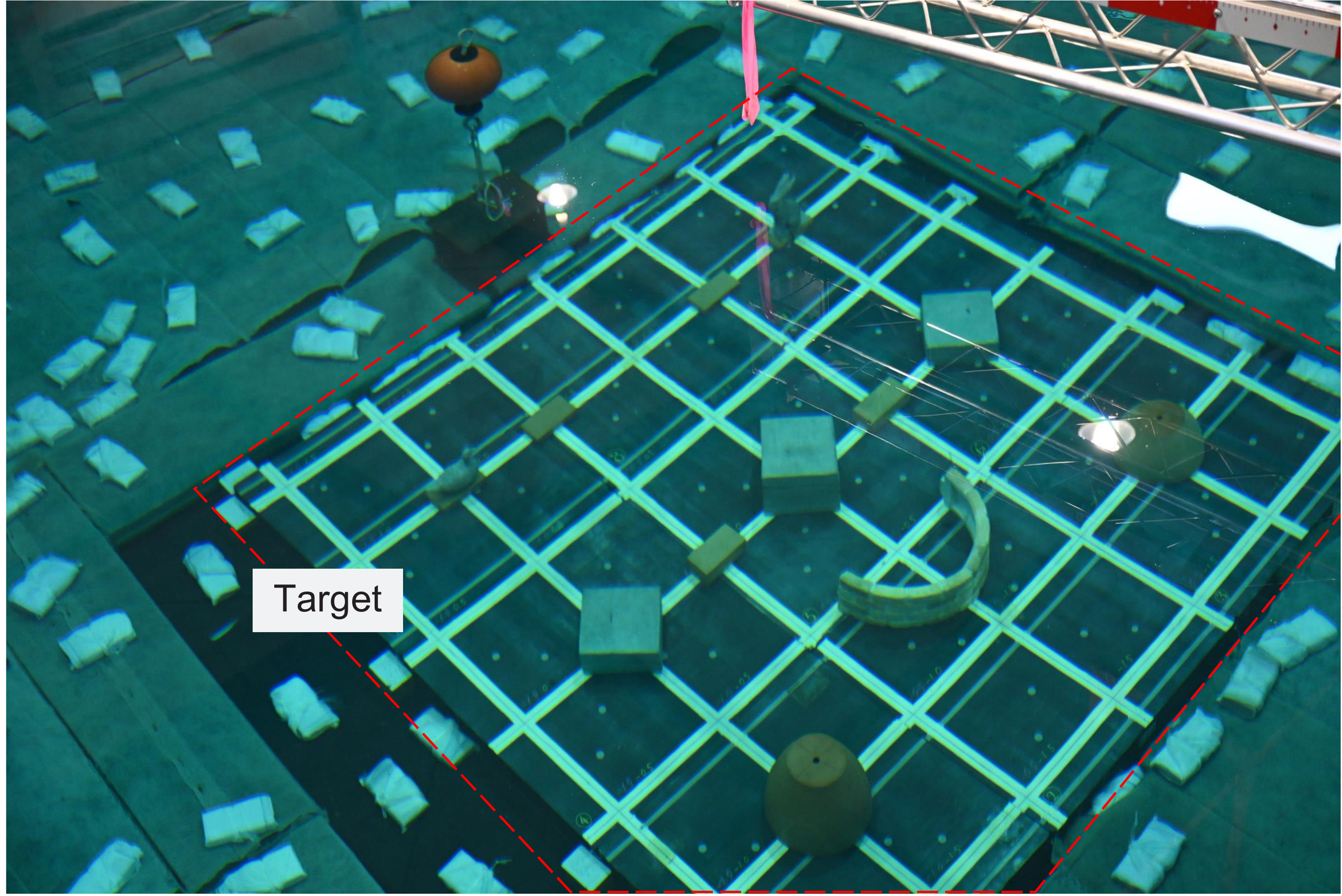}}
\caption{Water tank experiment. (a) Moving device for collecting data from different positions. (b) Target in the large scale water tank. } 
\label{fig:real}
\end{figure}

We collected a real dataset in a large scale water tank with Sound Metrics ARIS EXPLORER 3000 imaging sonar. The configuration of the objects was designed and then set up by a diver in the water tank. The ground truth of the objects was acquired using Agisoft Metashape in land. As shown in Fig.~\ref{fig:real}, a moving device was built to move the sonar to different positions. The sonar was mounted on a Sound Metrics AR2 rotator for roll rotations. To acquire the ground truth pose of the sonar in water tank coordinates, we calibrated the initial pose of the sonar by matching the edges of the real and synthetic images \cite{wang2022}. The rest of the poses were calculated from the control input of the moving device and the roll rotations. The ground truth depth labels were generated by inputting the sonar pose into the simulator.

The sonar with the rotator was moved by the device to 51 positions. For each position, the sonar performed a roll rotation from -35$^{\circ}$ to 35$^{\circ}$ and an acoustic video was recorded. Denoting the roll angle for the reference frame as $\alpha^{\circ}$, the source frame is taken at the roll angle of approximately $(\alpha+7)^{\circ}$. In total, we generated 1,493 pairs of reference and source images. The dataset was then separated randomly at an 8:2 ratio for training and test. Thus, there were 1,194 pairs for training and 299 pairs for test. The resolution in the range direction for the real images was downsampled by twice. All the heights and widths for the synthetic images and real images were 512$\times$128. 

\subsection{Training}

\begin{figure*}[t]
\centering
\subfloat[ \label{acs}]{\includegraphics[width=0.119\columnwidth]{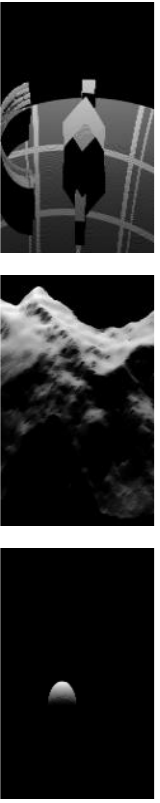}}\enskip
\subfloat[ \label{gt} Ground truth]{\includegraphics[width=0.38\columnwidth]{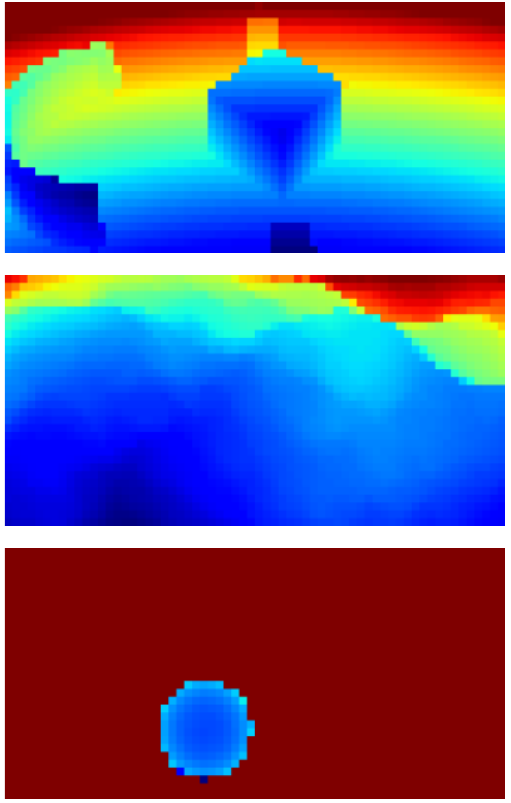}}\enskip
\subfloat[ \label{a2f} A2FNet \cite{Wangicra2021}]{\includegraphics[width=0.38\columnwidth]{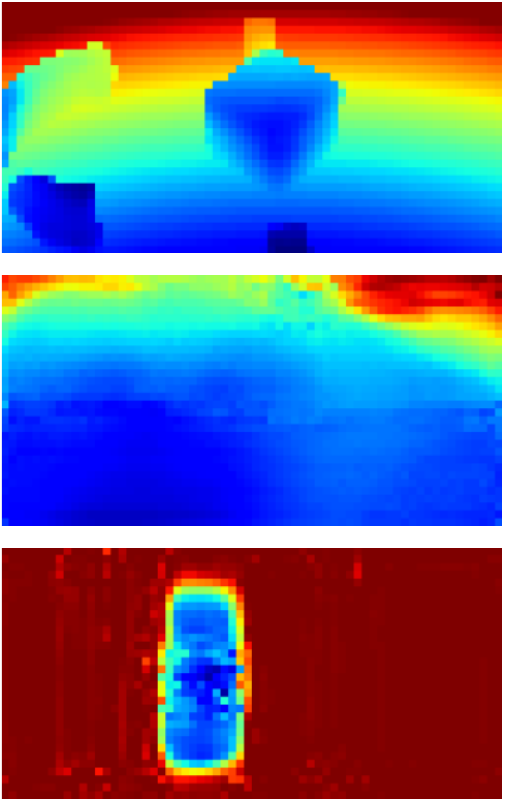}}\enskip
\subfloat[ \label{unet} ElevateNet \cite{DeBortoli2019}]{\includegraphics[width=0.38\columnwidth]{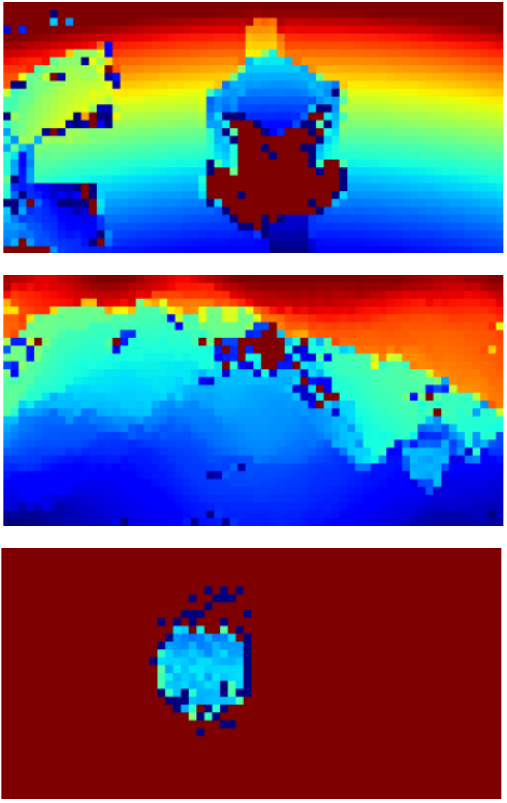}}\enskip
\subfloat[ \label{mvs} Ours ]{\includegraphics[width=0.38\columnwidth]{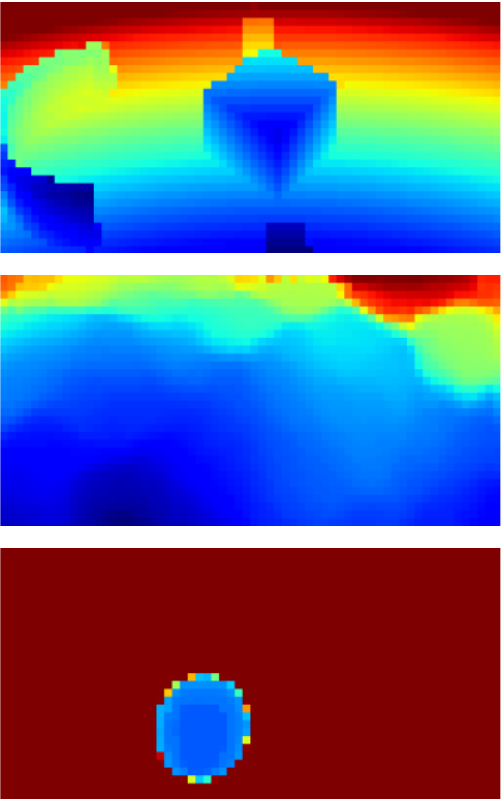}}
\caption{Visualization examples of simulation datasets. (a) Reference frames. Each row refers to a simulation dataset, which includes artificial (simulation), terrain, and sphere. The results are shown in depth maps. The color refers to depth values. Red and blue refer to the minimum and maximum, respectively. } 
\label{fig:sim_img}
\end{figure*}

\begin{figure*}[t]
\centering
\subfloat[ \label{realac}]{\includegraphics[width=0.118\columnwidth]{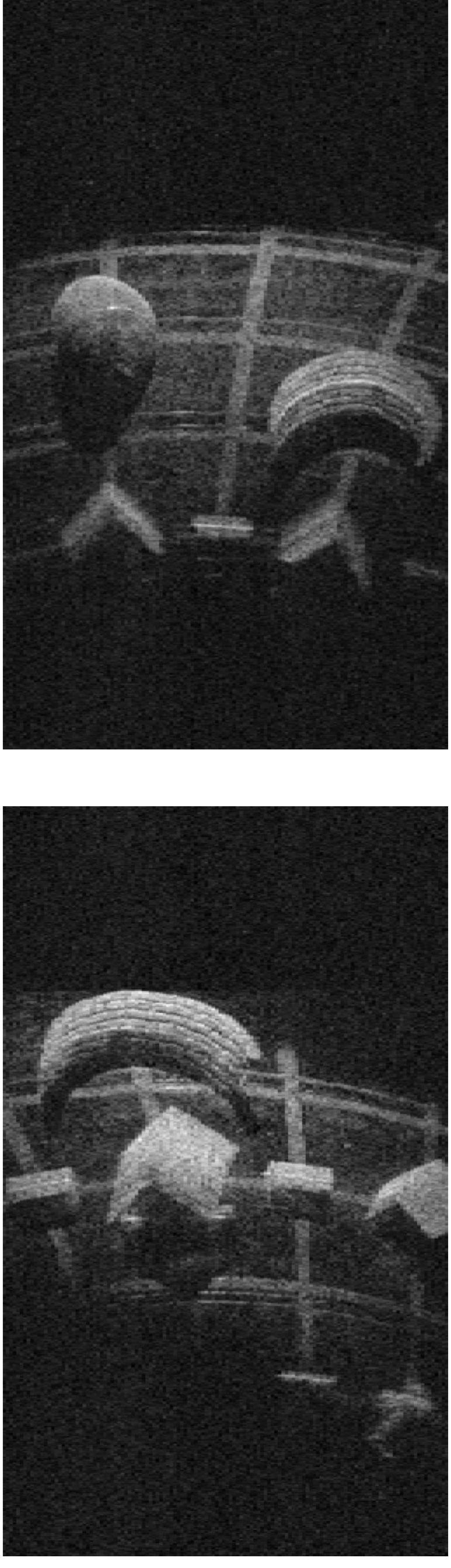}}\enskip
\subfloat[ \label{realgt} Ground truth]{\includegraphics[width=0.38\columnwidth]{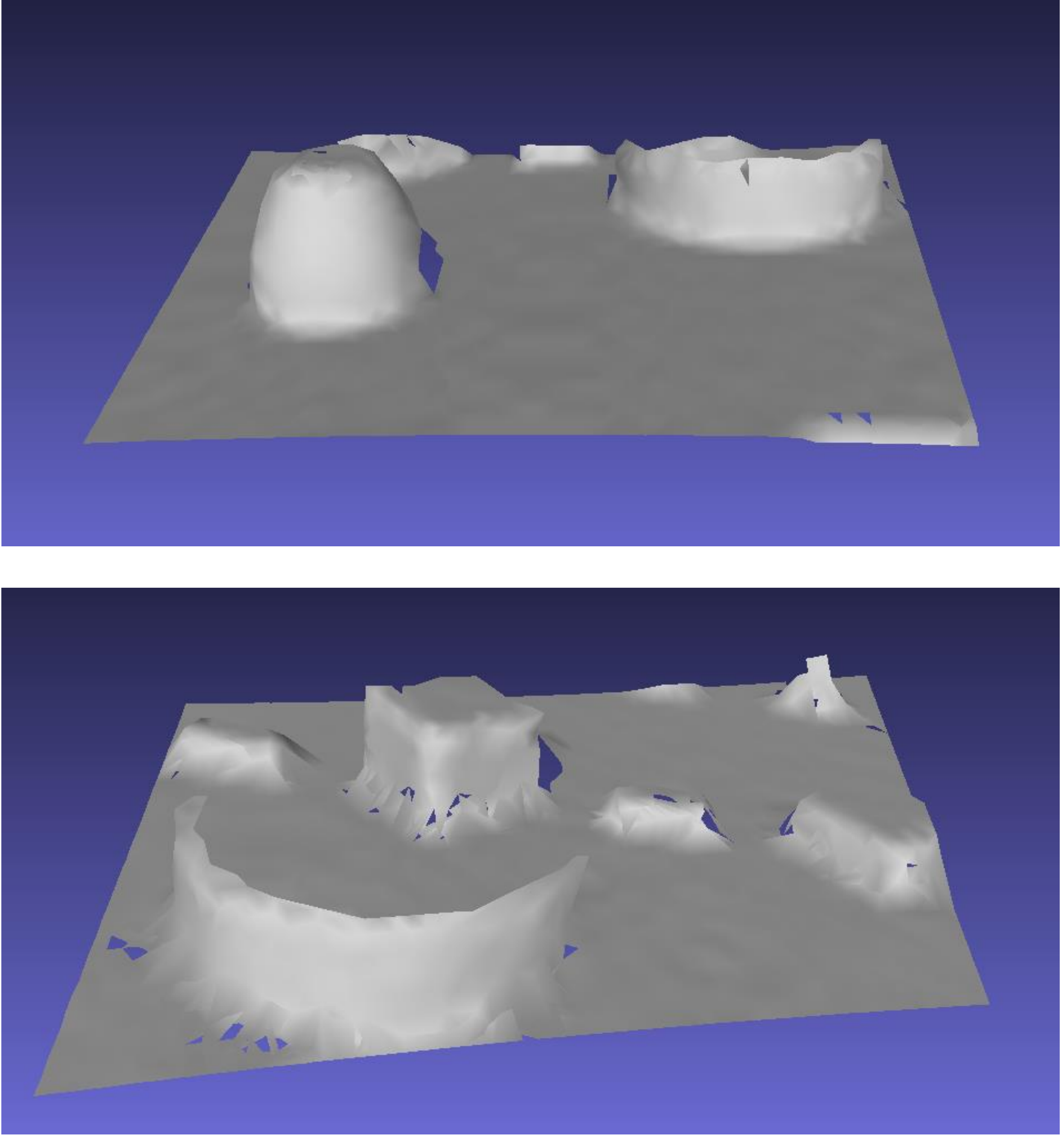}}\enskip
\subfloat[ \label{reala2f} A2FNet \cite{Wangicra2021}]{\includegraphics[width=0.38\columnwidth]{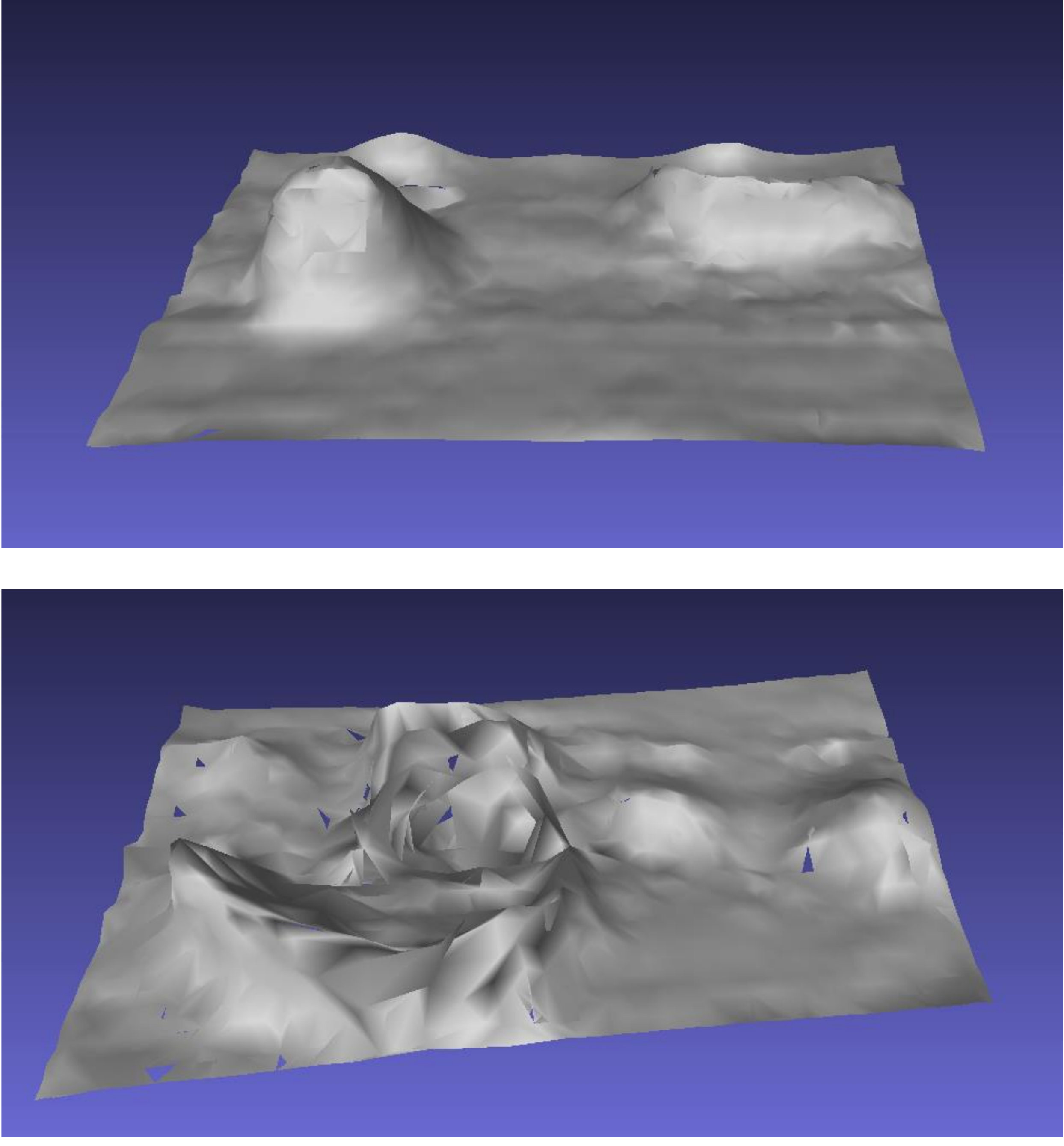}}\enskip
\subfloat[ \label{realunet} ElevateNet \cite{DeBortoli2019}]{\includegraphics[width=0.38\columnwidth]{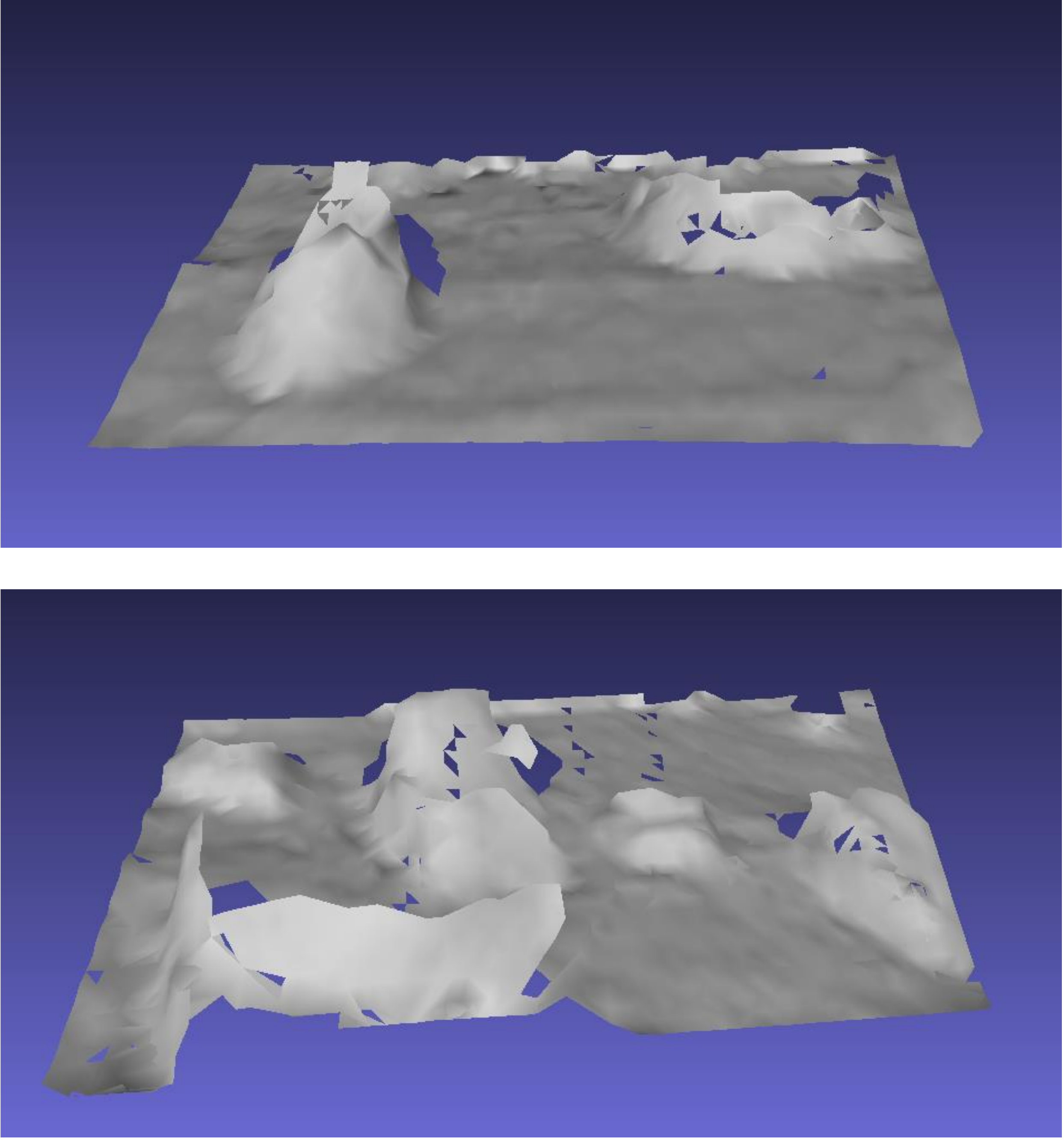}}\enskip
\subfloat[ \label{realmvs} Ours]{\includegraphics[width=0.38\columnwidth]{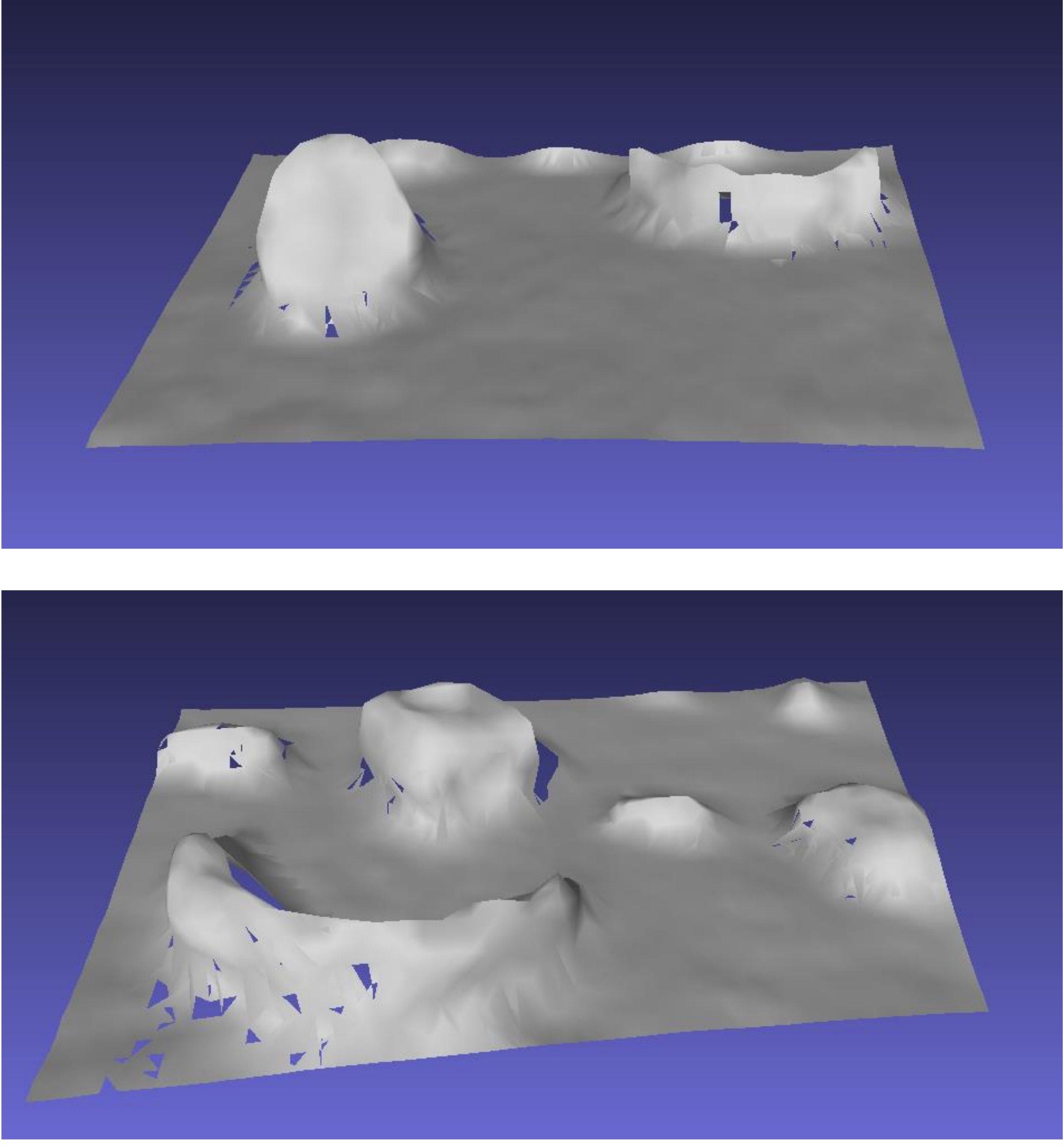}}
\caption{Visualization examples of real datasets. (a) Reference frames. We first transfer the estimation results to point clouds and then generate mesh models for comparison.} 
\label{fig:real_img}
\end{figure*}

The proposed method was implemented in PyTorch. The networks were trained and tested on NVIDIA GeForce RTX 3090 graphics card. For each dataset, we trained 50 epochs. The learning rate began at 0.001 and was reduced by half after 25 epochs. The batch size was set to 16 for training and test. Adam optimizer was used for optimization. $\lambda$ was set to 3.

\subsection{Metrics}

To evaluate the results, two metrics were used: the mean absolute error (MAE) for the front view depth map, and chamfer distance (CD) for the point cloud. 

\begin{equation}
    {\rm MAE} = \frac{\beta}{HW} \times \sum_{i=1}^{H}\sum_{j=1}^{W}|\hat{D}(i,j)-D(i,j)|,
\end{equation}
\begin{equation}
    {\rm CD} = \frac{\gamma}{S_1}\sum_{x \in S_1} \min_{y \in S_2}||x-y||_2^2+\frac{\gamma}{S_2}\sum_{y \in S_2} \min_{x \in S_1}||x-y||_2^2,
\end{equation}
where $\beta$ and $\gamma$ are scale parameters, which refer to 1,000 and 500, respectively. $S_1$ and $S_2$ refer to the two point sets.  

\subsection{Results}

To evaluate our method, we also compared our results with state-of-the-art methods, ElevateNet \cite{DeBortoli2019} and A2FNet \cite{Wangicra2021}. To implement ElevateNet, we chose UNet \cite{unet} as the backbone for training, with a sigmoid layer before the final output. Elevation maps were also generated using our simulator. The network was trained in a supervised manner using L1 loss. We also prepare a binary mask to segment the signal/non-signal region and the loss for the non-signal region was set to zero. During test time, the mask was multiplied to the network output. For A2FNet, the inverse pixel shuffle block was replaced by a four-layer CNN for feature extraction. We used the same hyper-parameters to train the baseline methods as our EPSSN.

The results of the simulation datasets are listed in Table~\ref{tab:Table1}. We show the average error and the standard deviation. All the metrics are the smaller the better. For artificial (simulation) and terrain datasets, our proposed method can generate more accurate and precise results compared to the baselines. For sphere datasets, the baseline methods using a single image can hardly deal with the problem; however, our proposed method can successfully estimate the 3D information. Figure~\ref{fig:sim_img} shows an example of the visualization of the results. For ElevateNet, although elevation maps were estimated, they were transferred to front view depth maps. The color refers to distance, where blue and red refer to the minimum and maximum distances, respectively. It can be seen from the results that our proposed method is closest to the ground truth. ElevateNet suffers from the incompleteness caused by non-bijective correspondence. For sphere datasets, A2FNet tends to estimate all the possible positions of the sphere; on the other hand, ElevateNet tends to estimate one of the possible results. By using two or more images, such ambiguity can be resolved and our proposed method can estimate the true results. Later, we will discuss how viewpoint numbers will influence our proposed method and the baseline methods. 

The results for the real dataset are listed in Table~\ref{tab:Table2}. Our proposed method is still far more better than the baseline methods. Examples of visualizing the results are shown in Fig.~\ref{fig:real_img}. We show the results of the mesh model for comparison. The mesh models were generated in MeshLab by first estimating normals and using ball pivoting. Clearly, the proposed method generated the 3D model with the best quality.  

\begin{table}[t]
\centering
\caption{Simulation results}\label{tab:Table1}
\scalebox{1.0}{
\begin{tabular}{|c|c|c|c|}
\hline
\multicolumn{4}{|c|}{Artificial (simulation)}\\ 
\hline 
\hline
& \multicolumn{1}{|c|}{A2FNet} & \multicolumn{1}{|c|}{ElevateNet}  &
\multicolumn{1}{|c|}{Ours}\\ 
\hline
CD [m]& 0.2707$\pm$0.1053 & 0.3263$\pm$0.1301   & \bf{0.1368$\pm$0.0473} \\
\hline
MAE [m]& 10.41 $\pm$ 3.09  &  -- & \bf{4.90$\pm$1.89}\\
\hline
 
\hline
\multicolumn{4}{|c|}{Terrain}\\ 
\hline 
\hline
& \multicolumn{1}{|c|}{A2FNet} & \multicolumn{1}{|c|}{ElevateNet}  &
\multicolumn{1}{|c|}{Ours}\\ 
\hline
CD [m]& 0.9962 $\pm$ 0.6155 & 0.5102 $\pm$ 0.2394   & \bf{0.2200$\pm$0.0837} \\
\hline
MAE [m]& 42.19$\pm$16.59  &  -- & \bf{14.07$\pm$4.67}\\
\hline

\hline
\multicolumn{4}{|c|}{Sphere        }\\ 
\hline 
\hline
& \multicolumn{1}{|c|}{A2FNet} & \multicolumn{1}{|c|}{ElevateNet}  &
\multicolumn{1}{|c|}{Ours}\\ 
\hline
CD [m]& 24.7811 $\pm$ 27.1205 &  9.6571 $\pm$ 9.4248  & \bf{1.4603$\pm$2.4521} \\
\hline
MAE [m]& 71.56$\pm$20.46 &  -- & \bf{9.60$\pm$5.87} \\
\hline
 
\end{tabular}

}
\end{table}

\begin{table}[t]
\centering
\caption{Real Experiment results}\label{tab:Table2}
\scalebox{1.0}{
\begin{tabular}{|c|c|c|c|}
\hline
\multicolumn{4}{|c|}{Artificial (real water tank)}\\ 
\hline 
\hline
& \multicolumn{1}{|c|}{A2FNet} & \multicolumn{1}{|c|}{ElevateNet}  &
\multicolumn{1}{|c|}{Ours}\\ 
\hline
CD [m]& 0.8677$\pm$0.2583 & 0.7605$\pm$0.2234   & \bf{0.2822$\pm$0.0879} \\
\hline
MAE [m]& 30.54 $\pm$ 7.07  &  -- & \bf{13.62$\pm$2.94}\\
\hline

\end{tabular}

}
\end{table}

\subsection{Discussions}
\subsubsection{Viewpoint Numbers}
The viewpoint numbers may influence the results. For a fair comparison, we also added viewpoints for ElevateNet and A2FNet by concatenating the reference and source images as inputs and modifying the channels of the networks. We chose the artificial and terrain datasets for training and test. The performance of ElevateNet seems unstable which significantly improved on terrain dataset but became worse on artificial dataset. This may because the 2D-3D problem is more obvious on artificial dataset. Our proposed method still outperforms the baseline. We also tied to increase the viewpoint number for EPSSN to three on terrain dataset, the CD was \textbf{0.1333$\pm$0.0573} m, which shows significant improvements. 

\begin{table}[t]
\centering
\caption{Viewpoint Numbers}\label{tab:Table_cat}
\scalebox{1.0}{
\begin{tabular}{|c|c|c|c|}
\hline
\multicolumn{4}{|c|}{Artificial (simulation)}\\ 
\hline 
\hline
& \multicolumn{1}{|c|}{A2FNet (two)} & \multicolumn{1}{|c|}{ElevateNet (two)}  &
\multicolumn{1}{|c|}{Ours}\\ 
\hline
CD [m]& 0.2707$\pm$0.1053
 & 0.3745$\pm$0.1597
   & \bf{0.1368$\pm$0.0473} \\
\hline
\multicolumn{4}{|c|}{Terrain}\\ 
\hline 
\hline
& \multicolumn{1}{|c|}{A2FNet (two)} & \multicolumn{1}{|c|}{ElevateNet (two)}  &
\multicolumn{1}{|c|}{Ours}\\ 
\hline
CD [m]& 0.9097$\pm$0.3439
 & 0.3541$\pm$0.1040
   & \bf{0.2200$\pm$0.0837} \\
\hline
 
\end{tabular}

}
\end{table}


\subsubsection{Regression on Elevation Map}

\begin{figure}[t]
\centering
{\includegraphics[width=1.0\columnwidth]{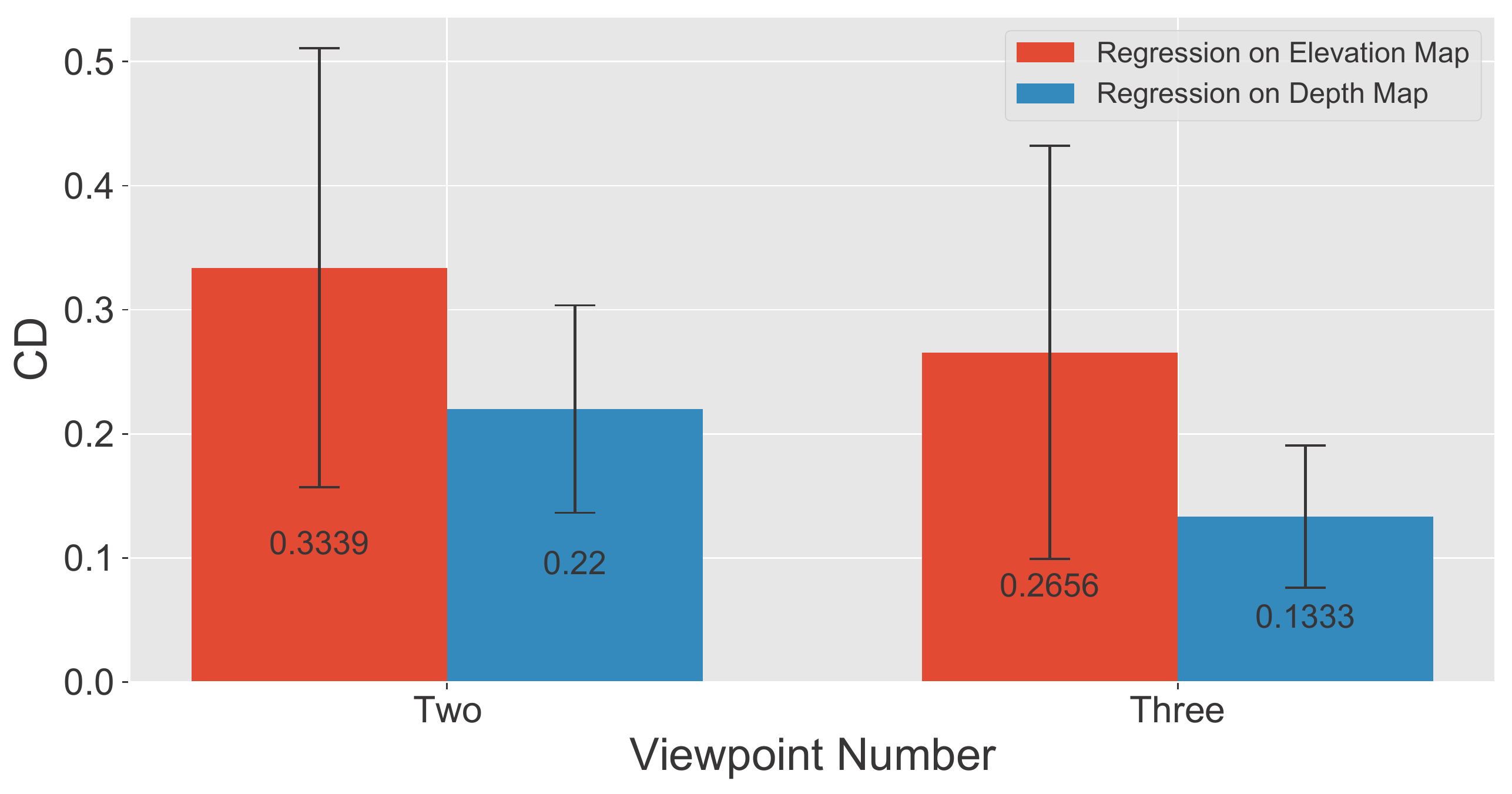}}
\caption{Ablation test results using the CD. Doing regression on the depth outperforms the regression on elevation. Increasing viewpoints may improve the performance of the proposed method. } 
\label{fig:ablation}
\end{figure}

Front depth generation from the cost volume and regression on the front depth map is one of the main contributions of this work. We tried to replace this part by generating elevation map and performing regression on it. 
Figure~\ref{fig:ablation} shows the performance of elevation map regression and depth map regression on CD. For the same number of viewpoints, depth map regression significantly outperforms elevation map regression. 

\subsubsection{Ablation on Elevation Plane Warping}

An ablation test was carried out on the elevation plane warping block when there were more viewpoints (three). By removing the warping process and simply adding the cost volumes from the reference frame and the source frames, the final result may be even worse than the case using two viewpoints with warping as shown in Table~\ref{tab:aba_warp}. In other words, the elevation plane warping successfully fuse the information from multiple frames with known motion information.  

\begin{table}[tb]
	\centering
	\caption{Ablation test on elevation plane warping}\label{tab:aba_warp}
	\scalebox{1.0}{
		\begin{tabular}{|c|c|c|}
			\hline
			\multicolumn{3}{|c|}{Terrain}\\ 
			\hline 
			\hline
			&  w / o warping (three) &
			w warping (three)\\ 
			\hline
			CD [m]& 0.2734$\pm$0.0819   & \bf{0.1333$\pm$0.0573} \\
			\hline
			MAE [m]& 16.26 $\pm$ 4.59  &  \bf{9.13$\pm$2.70}\\
			\hline
			
		\end{tabular}
	}
\end{table}

\subsubsection{Comparison with Occupancy Mapping}
We also compared the proposed method with a previous method using occupancy mapping \cite{Wangjoe}. We used two frames, one reference frame and one source frame, with known relative poses to generate the local map. With only two frames with a small roll angle difference, the results of occupancy mapping did not converge. On terrain dataset, the CD error for the test set was $46.83\pm 19.28$. 

\section{CONCLUSIONS}
In this study, we proposed a novel method for retrieving the missing 3D information in the acoustic images. By utilizing elevation plane warping to fuse the information from multiple frames and learning the pseudo front depth, our proposed method achieved state-of-the-art performance on both simulation and real datasets. It can also deal with severe ambiguous cases such as submerging sphere in water bodies. Our simulation datasets were made open source for future comparisons. The methods can be applied to underwater vehicles with two forward-looking sonars or a single forward-looking sonar with multiple views. 

Future work may include self-supervised network training and attempting various motions. Collecting datasets in real underwater environments is extremely difficult. In this study, the datasets were collected in a structured water tank where depth labels could be generated from our simulator. However, this is difficult to transfer to oceanic environments. Training the network without depth labels is worth investigating. Other methods, such as generative adversarial networks (GANs) may be valuable for mitigating the sim-to-real gap so that we can train the network based on synthetic data \cite{liu2021}. In this work, the motions between the reference frame and the source frames were pure roll rotation; other motions may also be tested in the future. 

\printbibliography


\addtolength{\textheight}{-12cm}   









\end{document}